\documentclass[letterpaper, 10 pt, conference]{ieeeconf}  

\IEEEoverridecommandlockouts                              

\usepackage{url}
\usepackage{graphics} 
\usepackage{graphicx}
\usepackage{epsfig} 
\usepackage{svg} 
\usepackage{subfigure}
\usepackage{times} 
\usepackage{amsmath} 
\usepackage{amssymb}  
\usepackage{algorithm}
\usepackage{multirow}
\usepackage{algpseudocode}
\usepackage{bm}
\usepackage{threeparttable}
\usepackage{booktabs}
\usepackage{xcolor}
\usepackage[verbose]{hyperref}

\title{\LARGE \bf
PI-WAN: A Physics-Informed Wind-Adaptive Network for Quadrotor Dynamics Prediction in Unknown Environments
}

\author{Mengyun Wang, Bo Wang, Yifeng Niu* and Chang Wang
\thanks{This work was supported in part by the National Natural Science Foundation of China under Grant 61876187, and in part by the Postgraduate Scientific Research Innovation Project of Hunan Province under Grant CX20240005. (\textit{Corresponding author: Yifeng Niu.})}
\thanks{Mengyun Wang, Bo Wang, Yifeng Niu and Chang Wang are with the College of Intelligence Science and Technology, National University of Defense Technology, Changsha 410073, China. {\tt \small e-mail: \{wangmengyun, wangbo\_, niuyifeng, wangchang07\}@nudt.edu.cn}.}%
}

\begin{document}

\maketitle
\thispagestyle{empty}
\pagestyle{empty}

\begin{abstract}
Accurate dynamics modeling is essential for quadrotors to achieve precise trajectory tracking in various applications. Traditional physical knowledge-driven modeling methods face substantial limitations in unknown environments characterized by variable payloads, wind disturbances, and external perturbations. On the other hand, data-driven modeling methods suffer from poor generalization when handling out-of-distribution (OoD) data, restricting their effectiveness in unknown scenarios. To address these challenges, we introduce the Physics-Informed Wind-Adaptive Network (PI-WAN), which combines knowledge-driven and data-driven modeling methods by embedding physical constraints directly into the training process for robust quadrotor dynamics learning. Specifically, PI-WAN employs a Temporal Convolutional Network (TCN) architecture that efficiently captures temporal dependencies from historical flight data, while a physics-informed loss function applies physical principles to improve model generalization and robustness across previously unseen conditions. By incorporating real-time prediction results into a model predictive control (MPC) framework, we achieve improvements in closed-loop tracking performance. Comprehensive simulations and real-world flight experiments demonstrate that our approach outperforms baseline methods in terms of prediction accuracy, tracking precision, and robustness to unknown environments. 

\end{abstract}

\section{Introduction}
\begin{figure}[htpb]
	\centering
	\includegraphics[scale=0.4]{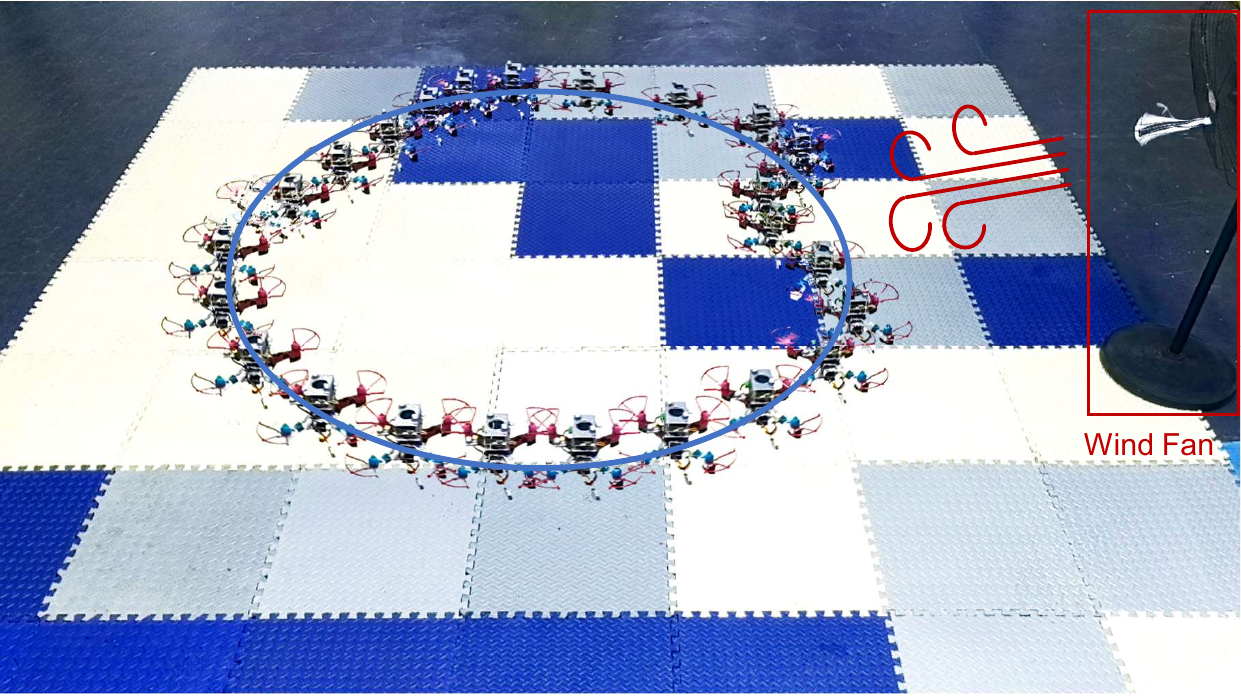}
	\caption{The quadrotor controlled by the proposed approach is tracking Circle trajectory in the presence of unknown external wind generated by a wind fan.}
	\label{real_flight}
\end{figure}

Accurate trajectory tracking of quadrotors in uncertain environments is crucial for accomplishing tasks such as search-and-rescue and autonomous transportation \cite{zuo2022unmanned}. Model-based control approaches \cite{wei2023mpcbased}, which leverage prior knowledge of system dynamics, have been widely adopted to enhance control performance \cite{spielberg2022neural}. However, real-world flight scenarios present significant challenges due to variable environmental factors such as payload changes \cite{ye2024oodcontrol}, wind disturbances \cite{oconnell2022neuralfly}, and ground effects \cite{shi2019neural}. 
Fig. \ref{real_flight} illustrates how wind disturbances can deflect the flight trajectory of a quadrotor in real-world conditions. 
These phenomena introduce nonlinear aerodynamics and disturbances that are difficult to characterize with analytical models, thereby significantly limiting the development and applicability of traditional model-based control methods \cite{sun2022comparative}.

Recent research has focused on developing precise models \cite{saviolo2023learning} of quadrotor dynamics to improve trajectory tracking accuracy. Simple linear drag models have demonstrated effectiveness in low-speed flight scenarios \cite{faessler2018differential}. However, high-speed and aggressive maneuvers typically induce complex nonlinear phenomena \cite{torrente2021datadriven}. Huang \emph{et al.} \cite{haomiaohuang2009aerodynamics} proposed modeling these complex aerodynamic forces using fundamental aerodynamic principles.  However, such knowledge-driven modeling methods require substantial expert knowledge and present significant challenges in practical applications \cite{bauersfeld2021neurobem}. 

Data-driven modeling methods \cite{zhu2024datadriven} have emerged as an alternative by leveraging historical flight data to improve dynamic modeling. Torrente \emph{et al.} \cite{torrente2021datadriven} employed Gaussian Processes (GPs) to represent residual dynamics during high-speed trajectory tracking. Although GPs offer advantages in modeling uncertainties, they suffer from significant computational limitations. Furthermore, their emphasis on unmodeled aerodynamics constrains their applicability in environments with unknown and variable conditions. 
To overcome these constraints, Saviolo \emph{et al.} \cite{saviolo2024active} proposed a neural network (NN) architecture capable of online parameter updating, enabling adaptation to dynamic and uncertain environments. This method demonstrated robust tracking performance under challenging conditions, including varying payloads, quadrotor configurations, and external wind disturbances.  However, these approaches primarily represent single-step dynamics without fully exploiting the temporal correlations inherent in sequential state transitions. 

Recurrent Neural Networks (RNNs) \cite{cho2023lowlevel} can model temporal sequences. However, their application in dynamics learning remains limited due to training instabilities caused by vanishing and exploding gradients. In contrast, Temporal Convolutional Networks (TCNs) implement causal and dilated convolutions to achieve stable and efficient model training. Lee \emph{et al.} \cite{lee2020learning} successfully applied TCNs to extract environmental characteristics for quadrupedal robots from historical state data. 
DroneDiffusion \cite{das2024dronediffusion} leveraged conditional diffusion models to learn quadrotor dynamics by processing state sequences. The learned model was integrated with an adaptive controller to generalize in complex, unseen scenarios. However, the computational complexity of diffusion models makes them impractical for online learning, especially for resource-constrained onboard deployment. 

Neural network-based dynamics often show limited generalization capabilities when predicting out-of-distribution (OoD) data. To address this issue, Ye \emph{et al.} \cite{ye2024oodcontrol} introduced random noises during training to improve model generalization on OoD data. Alternatively, integrating physical prior knowledge into learning frameworks \cite{sanyal2023rampnet} can reduce the dependence on training data while improving generalization to unseen data or new conditions. 
Saviolo \emph{et al.} \cite{saviolo2022physicsinspired} combined a sparse temporal convolution network (TCN) with a physics-inspired neural network (PINN) \cite{sanyal2023rampnet} to infer quadrotor dynamics from state histories. The physical information enhances predictive capabilities for OoD scenarios \cite{chrosniak2024deep}. However, this approach did not account for environmental disturbances during model learning. Moreover, a multi-layer perceptron (MLP) was used in the closed-loop tracking controller instead of the proposed PI-TCN structure to simplify optimization procedures. 

To address these limitations, we propose the Physics-Informed Wind-Adaptive Net (PI-WAN) to predict the dynamics of quadrotors under various wind conditions. 
Our main contributions are summarized as follows:

\begin{itemize}
    \item We combine knowledge-driven and data-driven modeling methods by embedding physical constraints into the training process of the PI-WAN. The TCN architecture efficiently captures temporal dependencies from historical flight data, while the physics-informed loss improves model generalization and robustness by applying physical knowledge to the learning framework. 
    \item The learned PI-WAN dynamics is employed as an estimator for external disturbances. The real-time prediction results are incorporated into the MPC framework to improve tracking performance in previously unseen trajectories, ensuring adaptability and robustness against various wind disturbances. 
    \item We conduct experiments in simulations and real-world scenarios. The proposed PI-WAN demonstrates superior prediction accuracy, tracking performance, and robustness under unknown wind conditions compared to baseline methods---including the nominal dynamics, TCN without physical constraints, and MLP architecture. 
\end{itemize}

The rest of the paper is organized as follows. Section \ref{method} introduces the PI-WAN and the MPC framework. Section \ref{exp} presents simulation and real-world experiments results. Finally, Section \ref{conclusion} concludes the key contributions and limitations and proposes the future directions. 

\begin{figure}[htpb]
	\centering
	\includegraphics[scale=0.4]{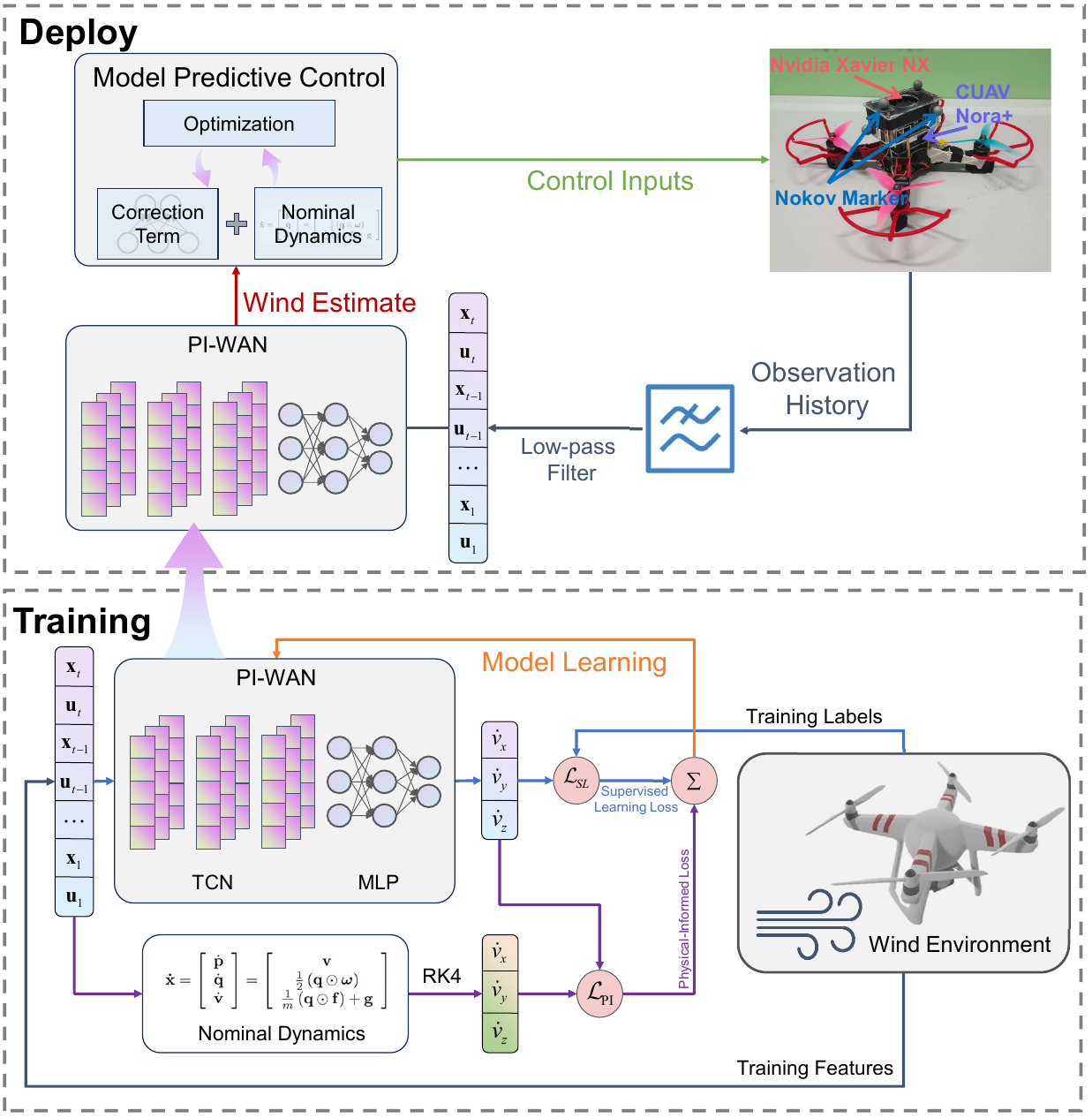}
	\caption{Architecture of the proposed approach. }
	\label{zong}
\end{figure}

\section{Methdology} \label{method}
\subsection{Nominal Dynamics}
The nominal dynamics of a quadrotor $\mathbf{\dot x} = f_{\text{nom}}\left(\mathbf{x}, \mathbf{u}\right)$ can be constructed by the state $\mathbf{x} = \left[\mathbf{p}^{\top}, \mathbf{q}^{\top}, \mathbf{v}^{\top}\right]^{\top} \in {\mathbb{R}}^{10}$ and the control input $\mathbf{u} = \left[ t_{\text{mn}}, \omega_x,  \omega_y, \omega_z\right]^{\top} \in {\mathbb{R}}^{4}$ as 

\begin{equation}
\label{nominal}
    \mathbf{\dot x} = 
\left[ \begin{array}{c}
\dot {\mathbf{p}}\\
\dot {\mathbf{q}}\\
\dot {\mathbf{v}}
\end{array} \right ]=
\left[ \begin{array}{cccc}
\mathbf{v}\\
\frac{1}{2}\left(\mathbf{q} \odot \bm{\omega}\right)\\
\frac{1}{m}\left(\mathbf{q} \odot \mathbf{f}\right) + \mathbf{g}
\end{array} \right ],
\end{equation}
where $\mathbf{p}, \mathbf{q}, \mathbf{v}$ are the position, quaternion, and the velocity in three-dimensional space. The control inputs contains mass-normalized thrust $t_{\text{mn}}$ and the body rates $\bm{\omega} = \left[ \omega_x,  \omega_y, \omega_z \right]^{\top}$. The thrust vector and the gravity vector in the inertia frame can be written as $\mathbf{f}=\left[0,0,t_{\text{mn}}\right]^{\top}$ and $\mathbf{g} = \left[0,0,g\right]^{\top}$. Given the state $\mathbf{x}_k$ and the control input $\mathbf{u}_k$ at timestamp $k$, the state at the next timestamp $\mathbf{x}_k$ can be computed by fourth-order Runge–Kutta integral method as 
\begin{equation}
    \mathbf{x}_{k+1} = \mathbf{x}_k + f_{\text{RK4}}\left( \mathbf{x}_k, \mathbf{u}_k, {\rm \Delta} T \right).
\end{equation}
However, the simplified nominal model derived from first principles fails to incorporate critical phenomena such as aerodynamic drag forces, rotor interactions, and environmental disturbances. 
These model inaccuracies inevitably result in degraded trajectory tracking performance. The primary objective of this paper is to develop a data-driven approach that leverages historical flight data to learn a more comprehensive and accurate dynamic model that surpasses the predictive capabilities of the nominal model. 

\subsection{Data Collection}
To develop a dynamics model with superior predictive capability under environmental disturbances, we systematically acquire flight data across a spectrum of wind velocities and trajectory patterns. 
Data collection is executed utilizing a MPC trajectory tracking framework that operates on a nominal dynamics model. 

\begin{figure}[htpb]
	\centering
	\includegraphics[scale=0.38]{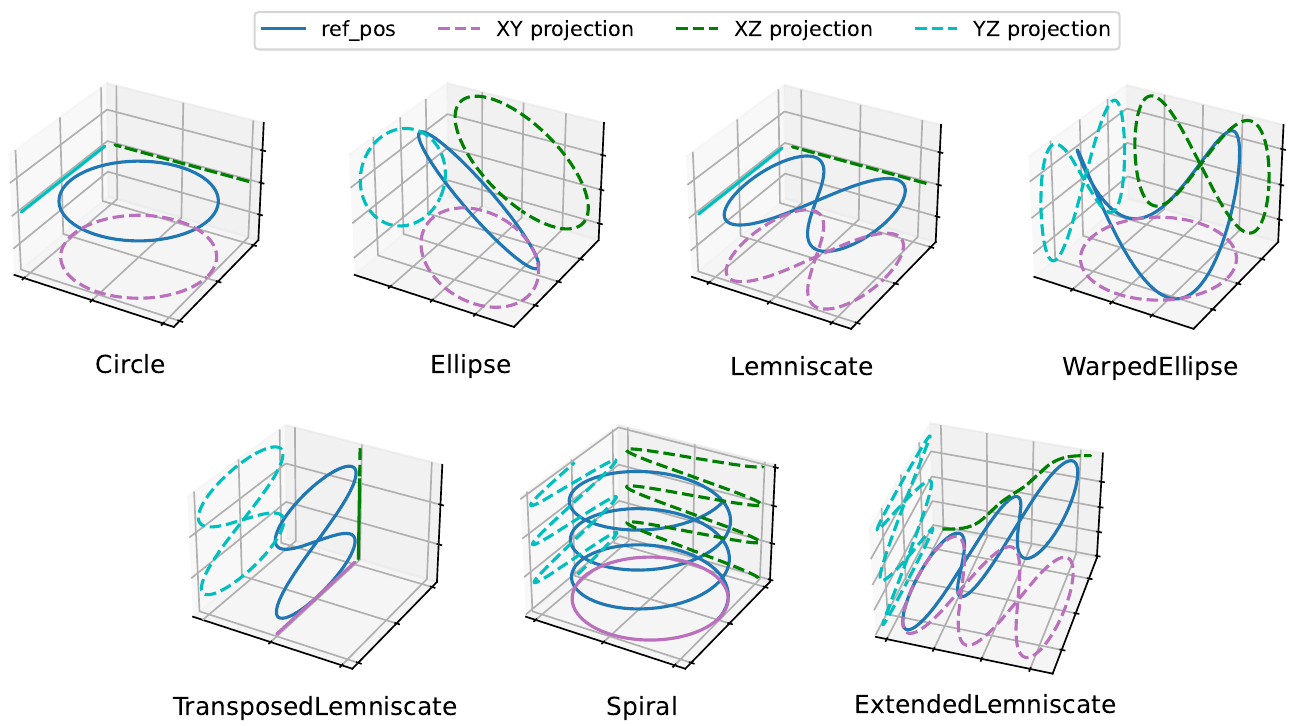}
	\caption{Different types of reference trajectories.}
	\label{traj}
\end{figure}

MPC governs system behavior to achieve optimal performance through iterative solution of a constrained optimization problem over a finite prediction horizon. The trajectory tracking problem is mathematically formulated as follows: 
\begin{equation}
\label{nom_mpc}
\begin{aligned}
    &\min_\mathbf{u}  \sum\limits_{k=0}^{H-1} \left[ \left( \mathbf{x}_k - \mathbf{x}_{k}^r \right)^{\top}\mathbf{Q} \left( \mathbf{x}_k - \mathbf{x}_{k}^r \right) + \mathbf{u}_k^{\top} \mathbf{R} \mathbf{u}_k \right]+\\
    &\quad \quad \left( \mathbf{x}_H - \mathbf{x}_{H}^r \right)^{\top}\mathbf{Q} \left( \mathbf{x}_H - \mathbf{x}_{H}^r \right).\\
    & \begin{array}{r@{\quad}l@{}l@{\quad}l}
    s.t. &\mathbf{x}_{k+1} = \mathbf{x}_k + f_{\text{RK4}}\left( \mathbf{x}_k, \mathbf{u}_k, {\rm \Delta} T \right),\\
         &\quad \mathbf{x}_0 = \mathbf{x}_{\rm{init}},\\
         &\mathbf{u}_{\min} \leq \mathbf{u}_k \leq \mathbf{u}_{\max},\\
\end{array}
\end{aligned}
\end{equation}
where $\mathbf{Q}$ and $\mathbf{R}$ represent semi-positive definite diagonal weighting matrices that penalize state tracking errors and control effort, respectively. The control inputs are bounded by constraints $\mathbf{u}_{\min}$ and $\mathbf{u}_{\max}$. 
The resulting constrained quadratic programming problem can be solved efficiently using sequential quadratic programming techniques. We leverage the CasADi framework \cite{andersson2019casadi} to deploy a multiple-shooting scheme that enhances computational efficiency and numerical stability. 

We design seven reference trajectories, including Circle, Ellipse, Lemniscate, TransposedLemniscate, Spiral, WarpedEllipse and ExtendedLemniscate. Each trajectory is executed for a duration of 20 seconds with temporal discretization at 0.02-second intervals, yielding 1000 data points per trajectory. We choose 5 of these reference trajectories to train the dynamics model, while the remaining two are reserved as unseen trajectories to test the generalization performance on OoD data. Environmental disturbances are introduced via directionally heterogeneous wind conditions, with independently configured velocity components along the x and y axes. 
Wind speeds are set to different values in the x and y directions. Throughout the flight, the state-action pairs of the quadrotor are recorded synchronously with their corresponding state derivatives.

\subsection{PI-WAN Training}
We design a Physics-Informed Wind-Adaptive Network (PI-WAN) to approximate the dynamics of the quadrotor under variable wind conditions. The network integrates state vectors and control input history to predict state derivatives. Specifically, the PI-WAN framework comprises two principal components: a Temporal Convolutional Network (TCN) that extracts temporal dependencies from sequential data, and a Multi-Layer Perceptron (MLP) that performs the nonlinear mapping to state derivatives. The TCN module implements dilated causal convolutions with residual connections, enabling it to efficiently process input sequences of arbitrary length while maintaining temporal coherence. This module encodes the historical trajectory into a fixed-dimension feature representation that captures relevant temporal dynamics across multiple time scales. 
Subsequently, this encoded representation serves as input to the MLP module, which approximates the complex nonlinear relationships governing the aerodynamic effects. The predicted dynamics represented by PI-WAN with parameters $\theta$ can be formally expressed as: 
\begin{equation}
    \mathbf{\dot x} = f_{\text{PI-WAN}}\left( \mathbf{X}_t, \mathbf{U}_t;\theta \right),
\end{equation}
where the inputs $\mathbf{X}_t = \left[ \mathbf{x}_{t-T}^{\top},  \mathbf{x}_{t-T+1}^{\top}, \cdots, \mathbf{x}_{t}^{\top}\right]^{\top}$ and $\mathbf{U}_t = \left[ \mathbf{u}_{t-T}^{\top},  \mathbf{u}_{t-T+1}^{\top}, \cdots, \mathbf{u}_{t}^{\top}\right]^{\top}$ denote the historical sequences of states and control inputs of length $T$ up to time $t$. The model learning problem is formulated as optimizing the parameters of PI-WAN to minimize the prediction error of the state derivatives. 
As observed in Eq. (\ref{nominal}), the derivative of position and quaternion can be computed directly from the state. Therefore, we focus exclusively on predicting the velocity derivatives as the network output. Furthermore, since position states do not directly influence the dynamics equations, they are omitted from the input feature. Moreover, the position state is not correlated with the dynamics. Consequently, the input tensor has dimensions $11\times T$, comprising quaternion $\mathbf{q}$, linear velocity $\mathbf{v}$, and control inputs $\mathbf{u}$. The output tensor has dimensions $3\times 1$, representing the velocity derivative $\mathbf{\dot v}$. 

Naive supervised learning suffers from poor generalization when faced with OoD data, resulting in significant prediction errors when models are deployed in new environments or real-world situations. Physics-Informed Neural Networks (PINNs) address this issue by embedding physical laws into neural networks through physics-informed constraints within the loss function. 
By integrating the differential equations of quadrotor dynamics into the learning process, PI-WAN effectively improves model robustness and generalization, especially when dealing with OoD data and conditions. 
During model training, the loss function can be written as 
\begin{equation}
    \mathcal{L} = \mathcal{L}_{\text{SL}} + \lambda \mathcal{L}_{\text{PI}},
\end{equation}
where $\mathcal{L}_{\text{SL}}$ represents the supervised learning loss based on labeled data, $\mathcal{L}_{\text{PI}}$ is the physics-informed loss, and $\lambda$ is a hyper-parameter balancing the contribution of each term.  $\mathcal{L}_{\text{SL}}$ is the mean squared error between training labels and the prediction output of PI-WAN. $\mathcal{L}_{\text{PI}}$ enforces physics-informed constraints by calculating the mean squared error between PI-WAN predictions and the nominal model outputs. The loss functions can be computed by
\begin{equation}
    \mathcal{L}_{\text{SL}} = \frac{1}{N_B} \sum_{i=1}^{N_B} \| \mathbf{\dot x}_{\text{labels}} - f_{\text{PI-WAN}} \left( \mathbf{X}_i, \mathbf{U}_i;\theta \right) \|,
\end{equation}
\begin{equation}
    \mathcal{L}_{\text{PI}} = \frac{1}{N_B} \sum_{j=1}^{N_B} \| f_{\text{RK4}}\left( \mathbf{x}_j, \mathbf{u}_j \right) - f_{\text{PI-WAN}} \left( \mathbf{X}_j, \mathbf{U}_j;\theta \right) \|,
\end{equation}
where $N_B$ is the numbers of a batch of training data, $\mathbf{\dot x}_{\text{labels}}$ is the training labels, $f_{\text{PI-WAN}} \left( \mathbf{X}_t, \mathbf{U}_t;\theta \right)$ is the prediction output of PI-WAN, and $f_{\text{RK4}}\left( \mathbf{x}_j, \mathbf{u}_j \right)$ is the state derivative computed by the nominal model. 

The calculation of $\mathcal{L}_{\text{PI}}$ requires random sampling from the state-input space to ensure comprehensive coverage. However, PI-WAN focuses on temporal sequence prediction rather than single-step dynamics, which inhibits direct random sampling. 
Previous work \cite{saviolo2022physicsinspired} utilized fixed pre-training samples for calculating PI loss, which limited the coverage of the state-input space. 
We periodically resample points at a predefined frequency during training. 
This dynamic sampling method ensures comprehensive coverage of physical information across the state-input space, resulting in more robust generalization properties. 

\subsection{Closed-Loop Control}
After establishing the MPC formulation for trajectory tracking \cite{salzmann2023realtime}, we integrate the trained PI-WAN to improve controller performance. 
While previous studies have incorporated learned models directly as hard constraints within MPC optimization frameworks, the computational complexity of neural networks often precludes real-time implementation. 
To enhance computational efficiency in optimization, Saviolo \emph{et al.} \cite{saviolo2022physicsinspired} demonstrated improved closed-loop control accuracy through physics-informed loss with simplified MLP architectures rather than computationally intensive TCNs. However, our approach requires inferring unknown wind disturbances from historical state sequences, a capability that surpasses MLP-based methods. 
We implement PI-WAN as an implicit compensator for the nominal model rather than as an explicit optimization constraint, thereby achieving a balance between predictive accuracy and computational efficiency. 

Assuming the difference between the nominal model's prediction and the PI-WAN's prediction results from environmental disturbances, we can characterize the environmental disturbance term as: 
\begin{equation}
f_w(\mathcal{H}_t) = \frac{1}{N_T} \sum_{i=1}^{N_T}(f_{\text{PI-WAN}}(\mathbf{x}_{t-i}, \mathbf{u}_{t-i}) - f_{\text{Nom}}(\mathbf{x}_{t-i}, \mathbf{u}_{t-i})),
\end{equation}
where $\mathcal{H}_t=\{\mathbf{x}_{t-N_T},\mathbf{u}_{t-N_T}, \cdots,\mathbf{x}_{t-1},\mathbf{u}_{t-1} \}$ represents the state-control history preceding time step $t$, and $N_T$ denotes the temporal window hyperparameter for disturbance estimation. Incorporating this disturbance term, we construct an enhanced dynamics model: 
\begin{equation}
f_{\text{corr}}\left(\mathbf{x}_t, \mathbf{u}_t, \mathcal{H}t\right) = f_{\text{Nom}}\left(\mathbf{x}_t, \mathbf{u}_t\right) + \boldsymbol{\Lambda}f_w(\mathcal{H}_t),
\end{equation}
where $\boldsymbol{\Lambda}$ is a disturbance adaptation matrix that dynamically adjusts the disturbance compensation intensity for each state component. 
The MPC formulation constrained by $f_{\text{corr}}$, as referenced in Eq. (\ref{nom_mpc}), can be expressed through the discrete update equation:
\begin{equation}
    \mathbf{x}_{k+1} = \mathbf{x}_k + \hat{f}_{\text{corr}}\left( \mathbf{x}_k, \mathbf{u}_k, \mathcal{H}_{t_k},{\rm \Delta} T \right),
\end{equation}
where $\hat{f}_{\text{corr}}$ is the Runge-Kutta numerical integration implementation of ${f}_{\text{corr}}$.

\begin{table*}
    \setlength\tabcolsep{5.5pt}
	\centering
	\caption{The prediction RMSE in different trajectories at various wind speeds.}
    \label{pre_rmse}
    \begin{tabular}{c||c|c|c|c|c||c|c}
    \toprule
    \textbf{Prediction RMSE [$m/s^2$]} & \multicolumn{5}{c||}{\textbf{Trained Trajectory}} & \multicolumn{2}{c}{\textbf{Unseen Trajectory}}\\
    \midrule
    Trajectory & Circle & Ellipse & Lemniscate & Lemniscate\_T & Spiral & WarpedEllipse & Lemniscate\_E\\
    \midrule
    Nom & 2.2720 & 2.6407 & 10.3416 & 11.9999 & 1.9896 & 2.3234 & 0.7551\\
    TCN & \textbf{0.0090} & \textbf{0.0108} & \textbf{0.0186} & \textbf{0.0208} & \textbf{0.0067} & 0.6804 & 0.4759\\
PI-MLP & 0.0115 & 0.0136 & 0.0283 & 0.0370 & 0.0112 & 1.3890 & 1.0782\\
PI-TCN & 0.0158 & 0.0212 & 0.0727 & 0.1026 & 0.0169 & 0.6776 & 0.2370\\
PI-WAN & 0.0230 & 0.0289 & 0.0667 & 0.1012 & 0.0196 & \textbf{0.4785} & \textbf{0.2007}\\
    \bottomrule
    \end{tabular}
    \color{black}
\end{table*}

\section{Experiments and Results} \label{exp}
In this section, we present a comprehensive evaluation of our proposed PI-WAN through simulations and experiments, aiming to answer the following questions:
1) How does the prediction performance of PI-WAN for quadrotor states under different wind speeds? Is it still robust on unseen reference trajectories and wind speeds that are not included in the training data?
2) How does integrating PI-WAN prediction results into closed-loop control improve trajectory tracking performance?
3) How does the overall performance of PI-WAN compared to other relevant baseline methods?
4) Can PI-WAN be deployed onboard on a real quadrotor?

\subsection{Environmental Setup}
\textbf{Implement.} We develop a custom wind-simulation environment using Airsim \cite{shah2018airsim} for evaluating our approach. We utilize two computers with 16 i9-9900K CPUs and an RTX 3080 GPU. One computer runs the high-fidelity Airsim simulation, while the other handles the training process. We set different wind speeds in the x-axis and y-axis directions to simulate the uncertainty of the real environment. Our PI-WAN is implemented using PyTorch. We implement the MPC framework based on CasADi \cite{andersson2019casadi}, and the weight matrix is set as $\mathbf{Q} = \text{diag}\left( 10, 10, 10, 5, 5, 5, 5, 1, 1, 1\right), \mathbf{R} = \text{diag} \left( 0.1, 0.2, 0.2, 0.2\right)$. 



\textbf{Baselines.} 
We compare the prediction performance of PI-WAN with four baselines: (1) PI-TCN, a TCN architecture trained in \cite{saviolo2022physicsinspired}; (2) PI-MLP, a multi-layer perceptron trained with the physics-informed loss; (3) TCN, a TCN trained without physics-informed loss; (4) Nom, the nominal model described in Eq. (\ref{nominal}).
Additionally, we evaluate the tracking performance of PI-WAN-MPC against four corresponding baselines: (1) PI-TCN-MPC; (2) PI-MLP-MPC; (3) TCN-MPC; (4) Nom-MPC. For each MPC variant, the corresponding dynamics model provides wind influence estimation, with the exception of Nom-MPC which relies exclusively on the nominal model without disturbance compensation. We use the root mean square error (RMSE) metric to evaluate the prediction and tracking performance.

\begin{figure}[htpb]
	\centering
	\includegraphics[scale=0.35]{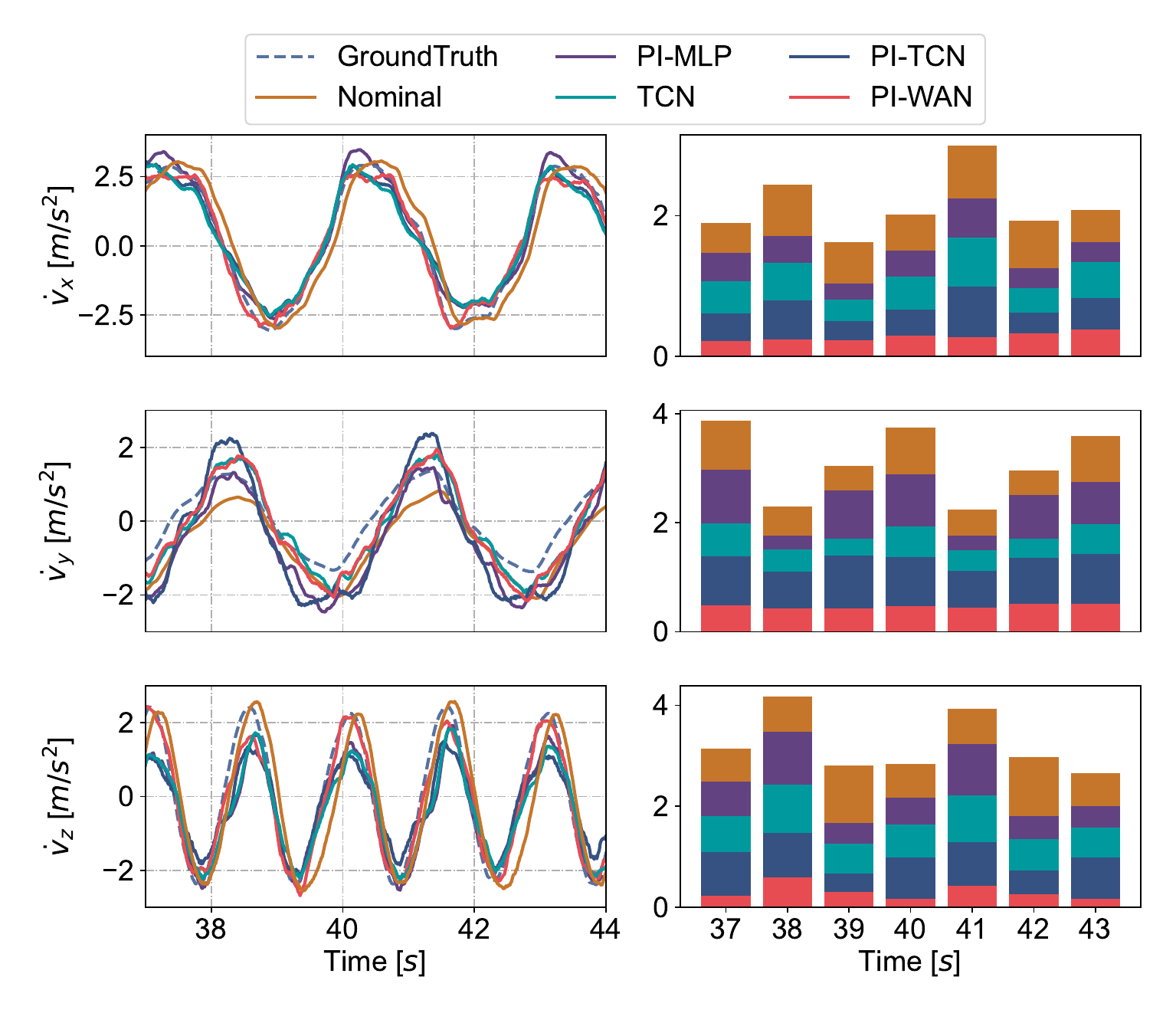}
	\caption{The prediction RMSE for the unseen reference trajectory. (\textbf{Left}: Prediction state derivatives. \textbf{Right}: RMSE between prediction and ground truth (stacked with no overlap).)}
	\label{pre_fig}
\end{figure}

\subsection{Predictive Performance}
We compare the predictive performance of PI-WAN with other baseline methods on training trajectories and previously unseen trajectories. 
Table \ref{pre_rmse} quantifies the predictive RMSE on different trajectories. Fig. \ref{pre_fig} visualizes the prediction results on the WarpedEllipse trajectory which was not encountered during training. The left plot shows the changes in predicted and actual linear acceleration, while the right bar chart presents the predictive RMSE across all baseline methods. 
The results show that the TCN, trained exclusively with supervised learning loss $\mathcal{L}_{\text{SL}}$ without the PI loss $\mathcal{L}_{\text{PI}}$, achieves the best performance on training trajectories but exhibits limited generalization in unseen trajectories. In contrast, physics-informed variants (PI-TCN and PI-WAN) that incorporate $\mathcal{L}_{\text{PI}}$ demonstrate superior performance on unseen test trajectories, despite sacrificing some prediction accuracy on the training trajectories. 
The proposed PI-WAN, which iteratively updates the sample points to calculate $\mathcal{L}_{\text{PI}}$, achieves higher tracking accuracy compared to PI-TCN. 
While PI-MLP shows competitive accuracy on training trajectories, its limitation in temporal feature extraction significantly impairs its ability to model environmental disturbances from sequential state data, resulting in substantially degraded performance on unseen trajectories. 

\begin{figure}[htpb]
	\centering
	\includegraphics[scale=0.58]{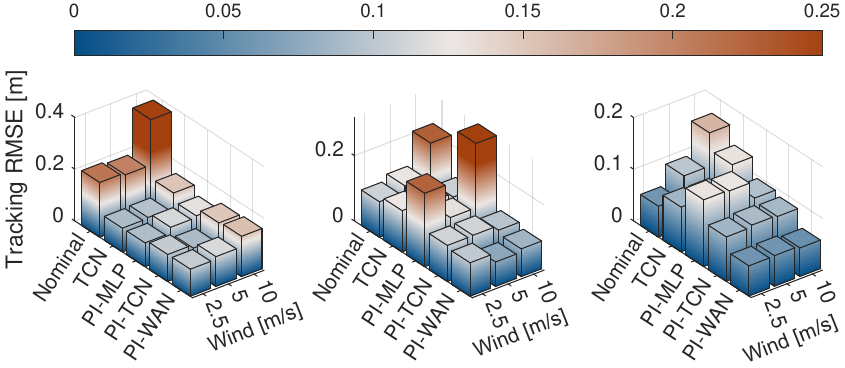}
	\caption{The heat map of tracking RMSE in different wind speeds and reference trajectories. (\textbf{Left}: Spiral trajectory. \textbf{Middle}: WarpedEllipse trajectory. \textbf{Right}: ExtendedLemniscate trajectory.)}
	\label{hot}
\end{figure}
\begin{table*}
    \setlength\tabcolsep{5.5pt}
	\centering
	\caption{The tracking RMSE in different trajectories at 5 $m/s$ wind in the x-axis.}
    \label{tracking_rmse}
    \begin{tabular}{c||c|c|c|c|c||c|c}
    \toprule
    \textbf{Tracking RMSE [$m$]} & \multicolumn{5}{c||}{\textbf{Trained Trajectory}} & \multicolumn{2}{c}{\textbf{Unseen Trajectory}}\\
    \midrule
    Trajectory & Circle & Ellipse & Lemniscate & Lemniscate\_T & Spiral & WarpedEllipse & Lemniscate\_E\\
    \midrule
    Nom-MPC & 0.1652 & 0.1581 & 0.7924 & 0.1931 & 0.2028 & 0.1270 & 0.0995\\
    TCN-MPC & 0.0940 & \textbf{0.1214} & \textbf{0.4722} & 0.1806 & 0.0950 & 0.1200 ($\downarrow$ 5.51\%) & 0.0852 ($\downarrow$ 14.37\%)\\
    PI-MLP-MPC & 0.1026 & 0.1595 & 0.4922 & \textbf{0.1801} & 0.1142 & 0.1204 ($\downarrow$ 5.20\%) & 0.1273 ($\uparrow$ 27.94\%)\\
    PI-TCN-MPC & 0.0792 & 0.1374 & 0.4772 & 0.1829 & \textbf{0.0907} & 0.1112 ($\downarrow$ 12.44\%) & 0.0893 ($\downarrow$ 10.25\%)\\
    PI-WAN-MPC & \textbf{0.0598} & 0.1355 & 0.4898 & 0.1886 & 0.1148 & \textbf{0.0747 ($\downarrow$ 41.18\%)}& \textbf{0.0592 ($\downarrow$ 40.50\%)}\\
    \bottomrule
    \end{tabular}
    \color{black}
\end{table*}

\begin{figure*}[htpb]
	\centering
	\includegraphics[scale=0.4]{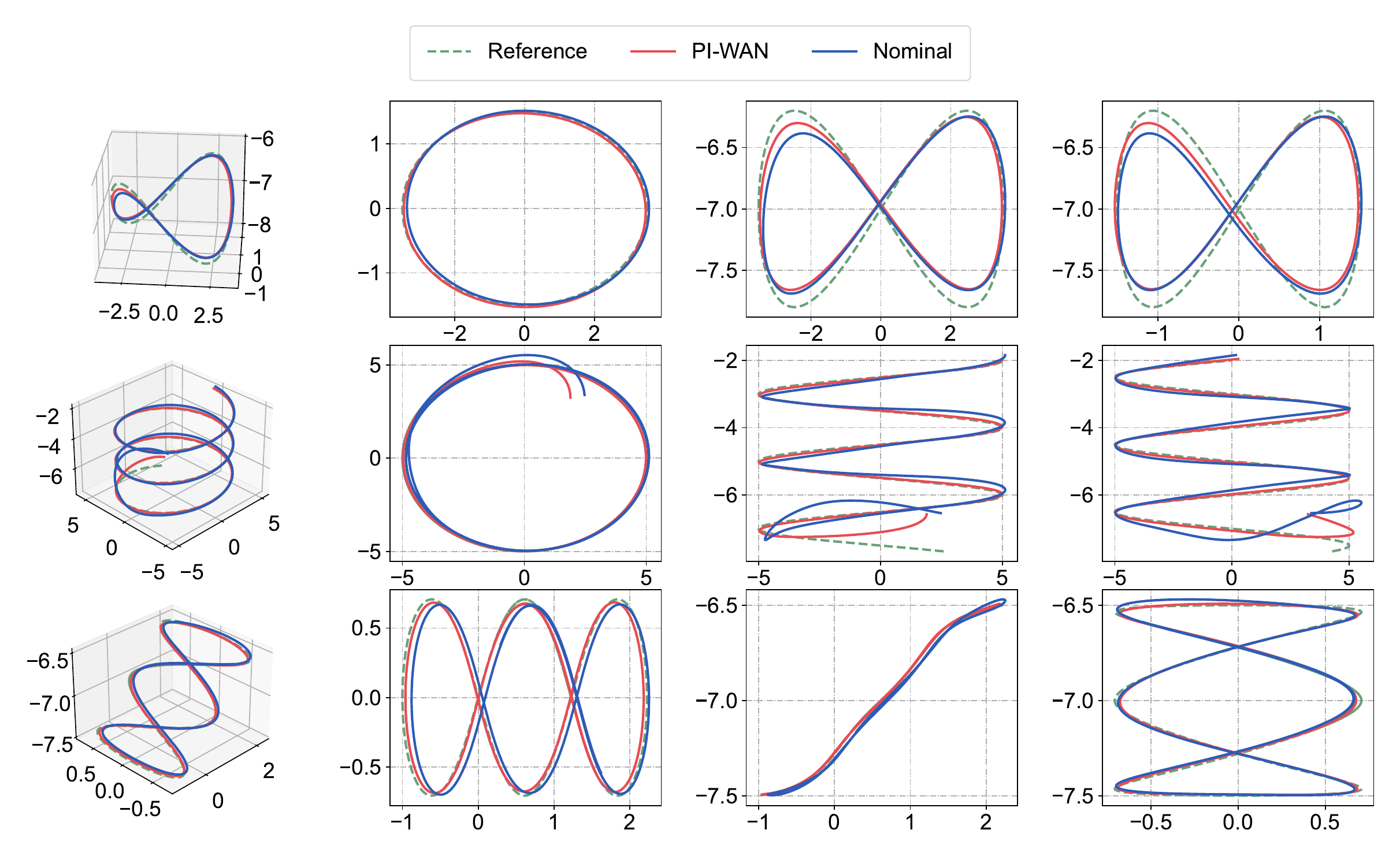}
	\caption{Visualization of trajectory tracking comparing PI-WAN and nominal approaches across various reference trajectories under wind disturbances. (\textbf{Left}: The 3D trajectories. \textbf{Right}: The 2D projections.)}
	\label{tracking}
\end{figure*}
\subsection{Closed-Loop Tracking Performance}
We integrate the state prediction PI-WAN model into the MPC framework to enhance trajectory tracking performance across diverse environmental conditions. Table \ref{tracking_rmse} compares the tracking RMSE across multiple trajectories. The results in the first row demonstrate that nominal MPC exhibits substantial tracking errors, primarily due to modeling inaccuracies caused by wind disturbances. In contrast, controllers augmented with learning-based prediction modules achieve significant improvements in tracking accuracy across both training and unseen trajectories. While multiple baseline controllers demonstrate comparable performance metrics on training trajectories, their capabilities diverge significantly when confronted with previously unseen trajectories. Only the physics-informed models (PI-TCN and PI-WAN) maintain robust performance enhancements. Other models even introduce additional decline in tracking accuracy. Figure \ref{tracking} visualizes the trajectory tracking results of nominal MPC and PI-WAN-MPC on Spiral, WarpedEllipse, and ExtendedLemniscate trajectories. 
All trials depicted in Figure \ref{tracking} are conducted under an x-axis wind speed of 5 $m/s$. 
The Spiral trajectory, included in the training dataset, demonstrates near-perfect alignment between the tracking and reference paths after correction by PI-WAN's predictive compensation. 
The previously unseen WarpedEllipse and ExtendedLemniscate trajectories also significantly improve when controlled by PI-WAN-MPC and achieve compensation for unknown disturbances, demonstrating its robustness and adaptability. 

Figs. \ref{hot} compares the tracking RMSE of 5 approaches under different wind speeds in Spiral, WarpedEllipse and ExtendedLemniscate trajectories. 
The nominal MPC has high tracking accuracy under minimal wind disturbance conditions, while the tracking error grows as wind velocity increases. The other controllers maintain tracking accuracy across the entire spectrum of wind conditions, attributable to their adaptive capacity to model and compensate for environmental disturbances. 
Controllers incorporating PI loss enable capturing physical principles, thereby maintaining excellent tracking performance on unseen trajectories.

\begin{figure}[htbp]
\centering
\includegraphics[width=7.5cm]{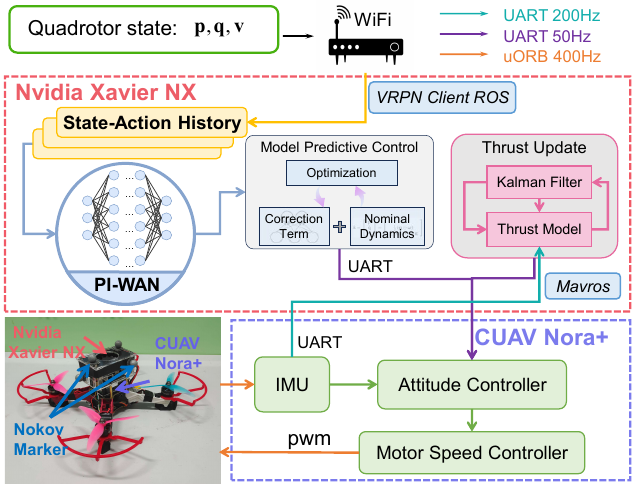}
\caption{The real-world experiment system. The quadrotor platform comprises the quadrotor frame, a CUAV Nora+ flight controller, and a Xavier NX onboard computer.}
\label{yingjian}
\end{figure}

\subsection{Real World Experiment}
We validate our controller augmented by PI-WAN using a real-world quadrotor. The physical experimental platform is illustrated in Fig. \ref{yingjian}. 
The quadrotor has a mass of 0.98$\, \rm{kg}$, with a maximum thrust-to-weight ratio of 4.5. The state of the quadrotor is provided by the Nokov motion capture system. 
Our algorithms are all implemented on the on-board computer Nvidia Xavier NX. Constrained by computational resources, we set the frequency of the MPC controller to 15 Hz. Disturbances estimation using PI-WAN is performed at 5 Hz. We map the mass-normalized thrust from the MPC to a throttle signal using the online thrust update module proposed in \cite{wang2024agile}. The attitude and throttle commands are sent to the flight control via the serial port. 
\begin{table}
	\centering
	\caption{The average tracking RMSE in the real-world experiments.}
    \label{RMSE_real}
    \begin{tabular}{c|c|c}
    \toprule
    \toprule
    \textbf{Approach} & \textbf{Nominal-MPC} & \textbf{PI-WAN-MPC}\\
    \midrule
    RMSE [$m$]& 0.2689 & 0.1927\\
    \bottomrule
    \bottomrule
    \end{tabular}
\end{table}

\begin{figure}[htbp]
\centering
\includegraphics[width=7.5cm]{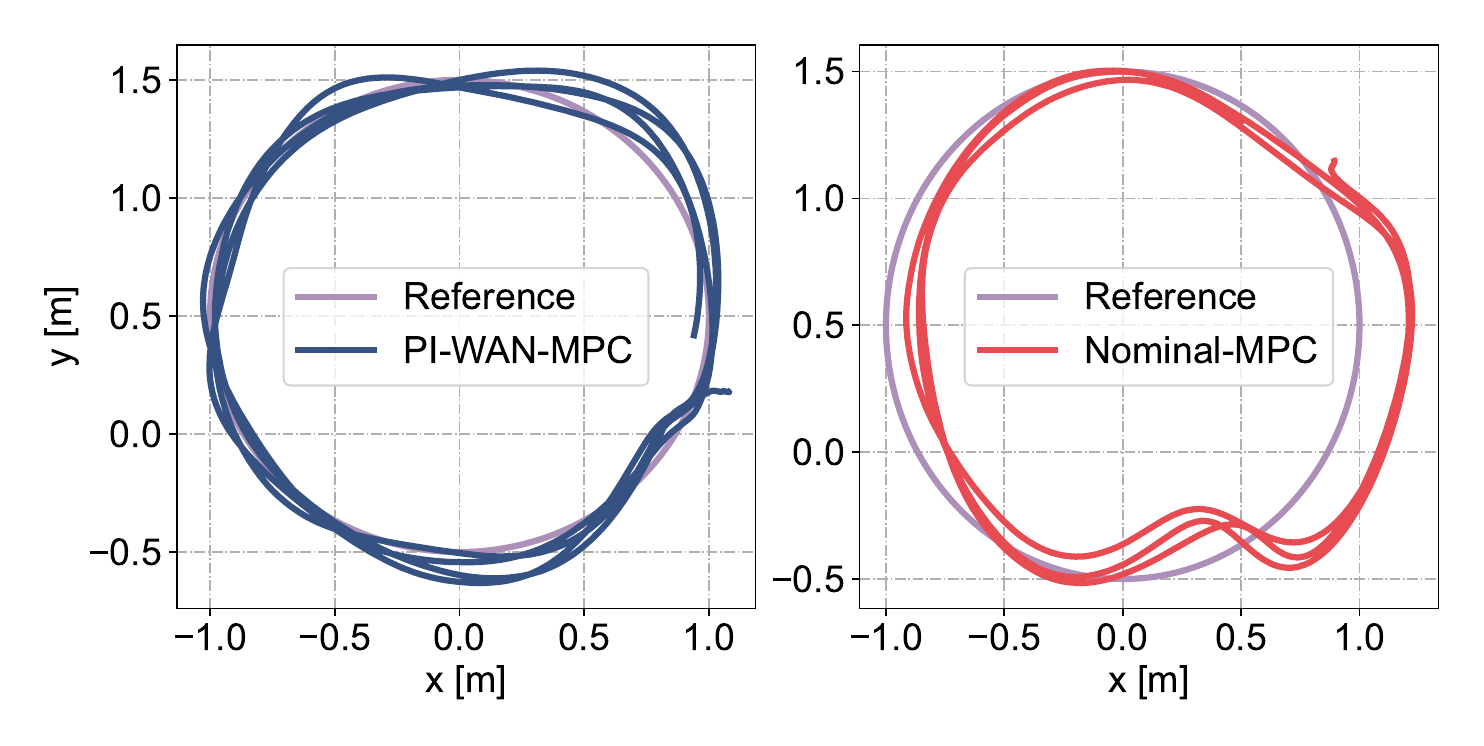}
\caption{The reference trajectory and the actual trajectory controlled by PI-WAN-MPC and Nominal MPC in real-world flights.}
\label{real_traj}
\end{figure}
We perform nominal MPC and PI-WAN-MPC under a circle trajectory with external wind generated by a fan. For fair comparison, the frequency and parameter settings of MPC are kept the same. Snapshots of the tracking process by PI-WAN-MPC are depicted in Fig. \ref{real_flight}.  
Each case is repeated at least three times, and we calculate the average tracking RMSE in Table \ref{RMSE_real}. 
PI-WAN has not been fine-tuned using real flight data. 
It can be seen that the quadrotor controlled by PI-WAN can accomplish the tracking task with small RMSE. However, the nominal MPC is subject to changing perturbations and has a large tracking error. The PI-WAN's ability to adapt to perturbations reduces the tracking error by 28.34\%. 
The consistent results with the simulation demonstrate the robustness of PI-WAN in the real world. 

\section{Conclusions and Future Work} \label{conclusion}
This paper introduces PI-WAN (Physics-Informed Wind-Adaptive Network), a novel framework that integrates knowledge-driven and data-driven modeling methods by embedding physical constraints directly into the training process for robust quadrotor dynamics learning. 
PI-WAN employs a TCN architecture with physics-informed loss functions, substantially enhancing model generalizability across previously unseen conditions. 
By incorporating real-time prediction results into a MPC framework, our approach enables real-time adaptation to external disturbances. Comprehensive experiments in both simulated and real-world scenarios demonstrate that PI-WAN outperforms baseline methods in prediction accuracy, trajectory tracking performance, and robustness to unknown environments. Future research directions include extending the framework to accommodate multiple environmental factors beyond wind effects and developing online learning capabilities to facilitate continuous adaptation to dynamic environmental conditions.









\bibliographystyle{IEEEtran} 
\bibliography{IEEEabrv, testbib} 

\begin{thebibliography}{10}
\providecommand{\url}[1]{#1}
\csname url@rmstyle\endcsname
\providecommand{\newblock}{\relax}
\providecommand{\bibinfo}[2]{#2}
\providecommand\BIBentrySTDinterwordspacing{\spaceskip=0pt\relax}
\providecommand\BIBentryALTinterwordstretchfactor{4}
\providecommand\BIBentryALTinterwordspacing{\spaceskip=\fontdimen2\font plus
\BIBentryALTinterwordstretchfactor\fontdimen3\font minus \fontdimen4\font\relax}
\providecommand\BIBforeignlanguage[2]{{%
\expandafter\ifx\csname l@#1\endcsname\relax
\typeout{** WARNING: IEEEtran.bst: No hyphenation pattern has been}%
\typeout{** loaded for the language `#1'. Using the pattern for}%
\typeout{** the default language instead.}%
\else
\language=\csname l@#1\endcsname
\fi
#2}}

\bibitem{zuo2022unmanned}
Z.~Zuo, C.~Liu, Q.-L. Han, and J.~Song, ``{Unmanned Aerial Vehicles}: Control methods and future challenges,'' \emph{IEEE/CAA J. Autom. Sin.}, vol.~9, no.~4, pp. 601--614, Apr. 2022.

\bibitem{wei2023mpcbased}
H.~Wei and Y.~Shi, ``{MPC}-based motion planning and control enables smarter and safer autonomous marine vehicles: Perspectives and a tutorial survey,'' \emph{IEEE/CAA J. Autom. Sin.}, vol.~10, no.~1, pp. 8--24, Jan. 2023.

\bibitem{spielberg2022neural}
N.~A. Spielberg, M.~Brown, and J.~C. Gerdes, ``Neural network model predictive motion control applied to automated driving with unknown friction,'' \emph{IEEE Trans. Control Syst. Technol.}, vol.~30, no.~5, pp. 1934--1945, Sept. 2022.

\bibitem{ye2024oodcontrol}
N.~Ye, Z.~Zeng, J.~Zhou, L.~Zhu, Y.~Duan, Y.~Wu, J.~Wu, H.~Zeng, Q.~Gu, X.~Wang, and C.~Zhou, ``{OoD-Control}: Generalizing control in unseen environments,'' \emph{IEEE Trans. Pattern Anal. Mach. Intell.}, vol.~46, no.~11, pp. 7421--7433, Nov. 2024.

\bibitem{oconnell2022neuralfly}
M.~O'Connell, G.~Shi, X.~Shi, K.~Azizzadenesheli, A.~Anandkumar, Y.~Yue, and S.-J. Chung, ``Neural-fly enables rapid learning for agile flight in strong winds,'' \emph{Sci. Robot.}, vol.~7, no.~66, p. eabm6597, May 2022.

\bibitem{shi2019neural}
G.~Shi, X.~Shi, M.~O'Connell, R.~Yu, K.~Azizzadenesheli, A.~Anandkumar, Y.~Yue, and S.-J. Chung, ``{Neural Lander}: Stable drone landing control using learned dynamics,'' in \emph{2019 International Conference on Robotics and Automation}.\hskip 1em plus 0.5em minus 0.4em\relax Montreal, QC, Canada: IEEE, May 2019, pp. 9784--9790.

\bibitem{sun2022comparative}
S.~Sun, A.~Romero, P.~Foehn, E.~Kaufmann, and D.~Scaramuzza, ``A comparative study of nonlinear {MPC} and differential-flatness-based control for quadrotor agile flight,'' \emph{IEEE Trans. Robot.}, vol.~38, no.~6, pp. 3357--3373, Dec. 2022.

\bibitem{saviolo2023learning}
A.~Saviolo and G.~Loianno, ``Learning quadrotor dynamics for precise, safe, and agile flight control,'' \emph{Annu. Rev. Control}, vol.~55, pp. 45--60, 2023.

\bibitem{faessler2018differential}
M.~Faessler, A.~Franchi, and D.~Scaramuzza, ``Differential flatness of quadrotor dynamics subject to rotor drag for accurate tracking of high-speed trajectories,'' \emph{IEEE Robot. Autom. Lett.}, vol.~3, no.~2, pp. 620--626, Apr. 2018.

\bibitem{torrente2021datadriven}
G.~Torrente, E.~Kaufmann, P.~Fohn, and D.~Scaramuzza, ``Data-driven {MPC} for quadrotors,'' \emph{IEEE Robot. Autom. Lett.}, vol.~6, no.~2, pp. 3769--3776, Apr. 2021.

\bibitem{haomiaohuang2009aerodynamics}
{Haomiao Huang}, G.~Hoffmann, S.~Waslander, and C.~Tomlin, ``Aerodynamics and control of autonomous quadrotor helicopters in aggressive maneuvering,'' in \emph{2009 IEEE International Conference on Robotics and Automation}.\hskip 1em plus 0.5em minus 0.4em\relax Kobe: IEEE, May 2009, pp. 3277--3282.

\bibitem{bauersfeld2021neurobem}
L.~Bauersfeld, E.~Kaufmann, P.~Foehn, S.~Sun, and D.~Scaramuzza, ``{NeuroBEM}: Hybrid aerodynamic quadrotor model,'' in \emph{Robotics: Science and Systems XVII}, July 2021.

\bibitem{zhu2024datadriven}
Y.~Zhu, H.~Cheng, and F.~Zhang, ``Data-driven dynamics modeling of miniature robotic blimps using neural odes with parameter auto-tuning,'' \emph{IEEE Robot. Autom. Lett.}, vol.~9, no.~12, pp. 10\,986--10\,993, Dec. 2024.

\bibitem{saviolo2024active}
A.~Saviolo, J.~Frey, A.~Rathod, M.~Diehl, and G.~Loianno, ``Active learning of discrete-time dynamics for uncertainty-aware model predictive control,'' \emph{IEEE Trans. Robot.}, vol.~40, pp. 1273--1291, 2024.

\bibitem{cho2023lowlevel}
J.-K. Cho, C.~Kim, M.~K.~M. Jaffar, M.~W. Otte, and S.-W. Kim, ``Low-level controller in response to changes in quadrotor dynamics,'' in \emph{2023 IEEE International Conference on Robotics and Automation}.\hskip 1em plus 0.5em minus 0.4em\relax London, United Kingdom: IEEE, May 2023, pp. 5317--5323.

\bibitem{lee2020learning}
J.~Lee, J.~Hwangbo, L.~Wellhausen, V.~Koltun, and M.~Hutter, ``Learning quadrupedal locomotion over challenging terrain,'' \emph{Sci. Robot.}, vol.~5, no.~47, p. eabc5986, Oct. 2020.

\bibitem{das2024dronediffusion}
A.~Das, R.~D. Yadav, S.~Sun, M.~Sun, S.~Kaski, and W.~Pan, ``{DroneDiffusion}: Robust quadrotor dynamics learning with diffusion models,'' 2024.

\bibitem{sanyal2023rampnet}
S.~Sanyal and K.~Roy, ``{RAMP-Net}: A robust adaptive {MPC} for quadrotors via physics-informed neural network,'' in \emph{2023 IEEE International Conference on Robotics and Automation (ICRA)}.\hskip 1em plus 0.5em minus 0.4em\relax London, United Kingdom: IEEE, May 2023, pp. 1019--1025.

\bibitem{saviolo2022physicsinspired}
A.~Saviolo, G.~Li, and G.~Loianno, ``Physics-inspired temporal learning of quadrotor dynamics for accurate model predictive trajectory tracking,'' \emph{IEEE Robot. Autom. Lett.}, vol.~7, no.~4, pp. 10\,256--10\,263, Oct. 2022.

\bibitem{chrosniak2024deep}
J.~Chrosniak, J.~Ning, and M.~Behl, ``{Deep Dynamics}: Vehicle dynamics modeling with a physics-informed neural network for autonomous racing,'' \emph{IEEE Robotics and Automation Letters}, vol.~9, no.~6, pp. 5292--5297, June 2024.

\bibitem{andersson2019casadi}
J.~A.~E. Andersson, J.~Gillis, G.~Horn, J.~B. Rawlings, and M.~Diehl, ``{CasADi}: A software framework for nonlinear optimization and optimal control,'' \emph{Math. Program. Comput.}, vol.~11, no.~1, pp. 1--36, Mar. 2019.

\bibitem{salzmann2023realtime}
T.~Salzmann, E.~Kaufmann, J.~Arrizabalaga, M.~Pavone, D.~Scaramuzza, and M.~Ryll, ``Real-time neural-{MPC}: Deep learning model predictive control for quadrotors and agile robotic platforms,'' \emph{Ieee Robot. Autom. Lett.}, pp. 1--8, 2023.

\bibitem{shah2018airsim}
S.~Shah, D.~Dey, C.~Lovett, and A.~Kapoor, ``{AirSim}: High-fidelity visual and physical simulation for autonomous vehicles,'' in \emph{Field and Service Robotics}, ser. Springer Proceedings in Advanced Robotics.\hskip 1em plus 0.5em minus 0.4em\relax Cham: Springer International Publishing, 2018, pp. 621--635.

\bibitem{wang2024agile}
M.~Wang, S.~Jia, Y.~Niu, Y.~Liu, C.~Yan, and C.~Wang, ``Agile flights through a moving narrow gap for quadrotors using adaptive curriculum learning,'' \emph{IEEE Trans. Intell. Veh.}, vol.~9, no.~11, pp. 6936--6949, Nov. 2024.

\end{thebibliography}

\end{document}